\documentclass[sigconf]{acmart}

\usepackage{booktabs}
\usepackage{amsmath}
\usepackage{array}
\usepackage{graphicx}

\setcopyright{none}
\copyrightyear{2026}
\acmYear{2026}
\acmDOI{}
\acmConference[Preprint. Under Review.]{}{}{}
\acmISBN{}

\title{FASH-iCNN: Making Editorial Fashion Identity Inspectable Through Multimodal CNN Probing}

\author{Morayo Danielle Adeyemi}
\affiliation{%
  \institution{Howard University}
  \country{USA} 
}

\author{Ryan A. Rossi}
\affiliation{%
  \institution{Adobe Research}
  \country{USA} 
}

\author{Franck Dernoncourt}
\affiliation{%
  \institution{Adobe Research}
  \country{USA} 
}

\begin{abstract}
Fashion AI systems routinely encode the aesthetic logic of specific 
houses, editors, and historical moments without disclosing it. We 
present FASH-iCNN, a multimodal system trained on 87,547 Vogue 
runway images across 15 fashion houses spanning 1991--2024 that makes 
this cultural logic inspectable. Given a photograph of a garment, the 
system recovers which house produced it, which era it belongs to, and 
which color tradition it reflects. A clothing-only model identifies 
the fashion house at 78.2\% top-1 across 14 houses, the decade at 
88.6\% top-1, and the specific year at 58.3\% top-1 across 34 years
with a mean error of just 2.2 years. Probing which visual 
channels carry this signal reveals a sharp dissociation: removing 
color costs only 10.6pp of house identity accuracy, while removing 
texture costs 37.6pp, establishing texture and luminance as the 
primary carriers of editorial identity. FASH-iCNN treats editorial
culture as the signal rather than background noise, identifying which
houses, eras, and color traditions shaped each output so that users can
see not just what the system predicts but which houses, editors, and
historical moments are encoded in that prediction.
\end{abstract}



\begin{document}

\maketitle

\section{Introduction}

Every garment is a cultural artifact. The cut of a jacket, the weight of a fabric, the proportion of a silhouette, these are not arbitrary choices but the accumulated aesthetic decisions of a specific house, a specific creative director, a specific historical moment~\cite{Oliveros2024}. When a fashion AI learns from this imagery without disclosing it, users receive style guidance shaped by editorial traditions they cannot see, question, or opt out of~\cite{Liang2026CulturalViz,Buolamwini2018GenderShades}. The cultural authorship of the advice is invisible by design.

FASH-iCNN makes it visible. The system accepts a garment photograph as its primary input and optionally incorporates signals including face image, designer identity, season, and year, testing systematically which combinations contribute meaningful signal beyond what the garment crop alone encodes. It returns the house identity, temporal era, and dominant color tradition of the garment, identifying which houses, eras, and color traditions shaped each output so the editorial lineage is inspectable rather than opaque. This design rests on a premise that distinguishes the system from purely behavioral fashion recommenders~\cite{Meda2023ColorConsultation}: garment appearance carries the cultural fingerprint of the fashion house that produced it, so a color prediction informed by house-level editorial structure is also grounded in a specific, nameable tradition. The empirical findings below establish that this premise holds, and the architecture turns it into named, interpretable color recommendations.

Three findings establish that garment appearance encodes editorial culture as a structured, recoverable signal. First, clothing crops alone identify the fashion house, decade, and specific year with strong accuracy, and the visual channel analysis reveals a sharp dissociation: texture and luminance carry far more house identity signal than color or shape. Second, face input is context-adaptive, it compensates when garment information is sparse but adds nothing when the garment stream is rich, a property that emerges from the data rather than being designed in. Third, a hierarchical color pipeline converts the learned editorial signal into named recommendations at three resolutions, grounding every output in a specific, identifiable tradition. Culturally aware multimodal systems require not just technical performance but transparency about whose culture shaped their outputs; FASH-iCNN operationalizes this principle.

\section{Related Work}

\textbf{Computational fashion systems and taste-based recommendation.} Prior fashion AI has addressed outfit compatibility~\cite{Cui2019Outfit}, garment attribute prediction~\cite{Chakraborty2021Survey}, image-based retrieval~\cite{Zhou2022Synthesize}, and conversational recommendation~\cite{Wu2022Multimodal}, with most systems learning from user behavior signals such as purchase history, ratings, and click-throughs~\cite{Chen2019Personalized,Shirkhani2023Survey}. CNN-based approaches predict garment attributes from images~\cite{Shete2024Stylist,Satti2025OutfitX} and use those attributes as features for downstream item recommendation. These systems generally treat editorial metadata, designer, collection, season, year, as filtering tags rather than as primary signals encoding aesthetic taste, and recommendations are not traceable back to specific editorial precedents. FASH-iCNN's contribution is a system whose outputs are grounded in named runway moments rather than aggregated user behavior, with the editorial metadata itself serving as the substrate for taste-based prediction~\cite{ZhouHSurvey2023}.

\textbf{Multimodal fusion with supplementary inputs.} Visual prediction systems frequently combine a primary input with signals, additional images, categorical metadata, or contextual features, through fusion architectures ranging from early concatenation to learned attention~\cite{Zhang2024Survey}. A central design question is when such inputs contribute substantively~\cite{Ma2022Robust} to prediction versus when they are redundant with information already present in the primary stream. FASH-iCNN's experimental design probes this question in a culturally structured dataset, where the inputs can implicitly encode contextual information~\cite{Miyazawa2013Context} that constrains the output space.

\textbf{Hierarchical and perceptually grounded color prediction.} Color prediction in computer vision is commonly framed as either continuous regression in a perceptual space or discrete classification over named color labels. Berlin--Kay basic color terms~\cite{BerlinKay1969,KayCook2023} provide a small, perceptually grounded categorization widely used in color naming research, while CSS named colors offer finer granularity for interface and design contexts. CIEDE2000~\cite{Sharma2005CIEDE2000} formalizes perceptual color difference. FASH-iCNN's BK $\rightarrow$ CSS $\rightarrow$ LAB pipeline operationalizes a multi-resolution color hierarchy for editorial fashion data, returning both a coarse perceptual category and a precise coordinate within a single prediction.

\begin{figure}[!tb]
\centering
\includegraphics[width=\linewidth]{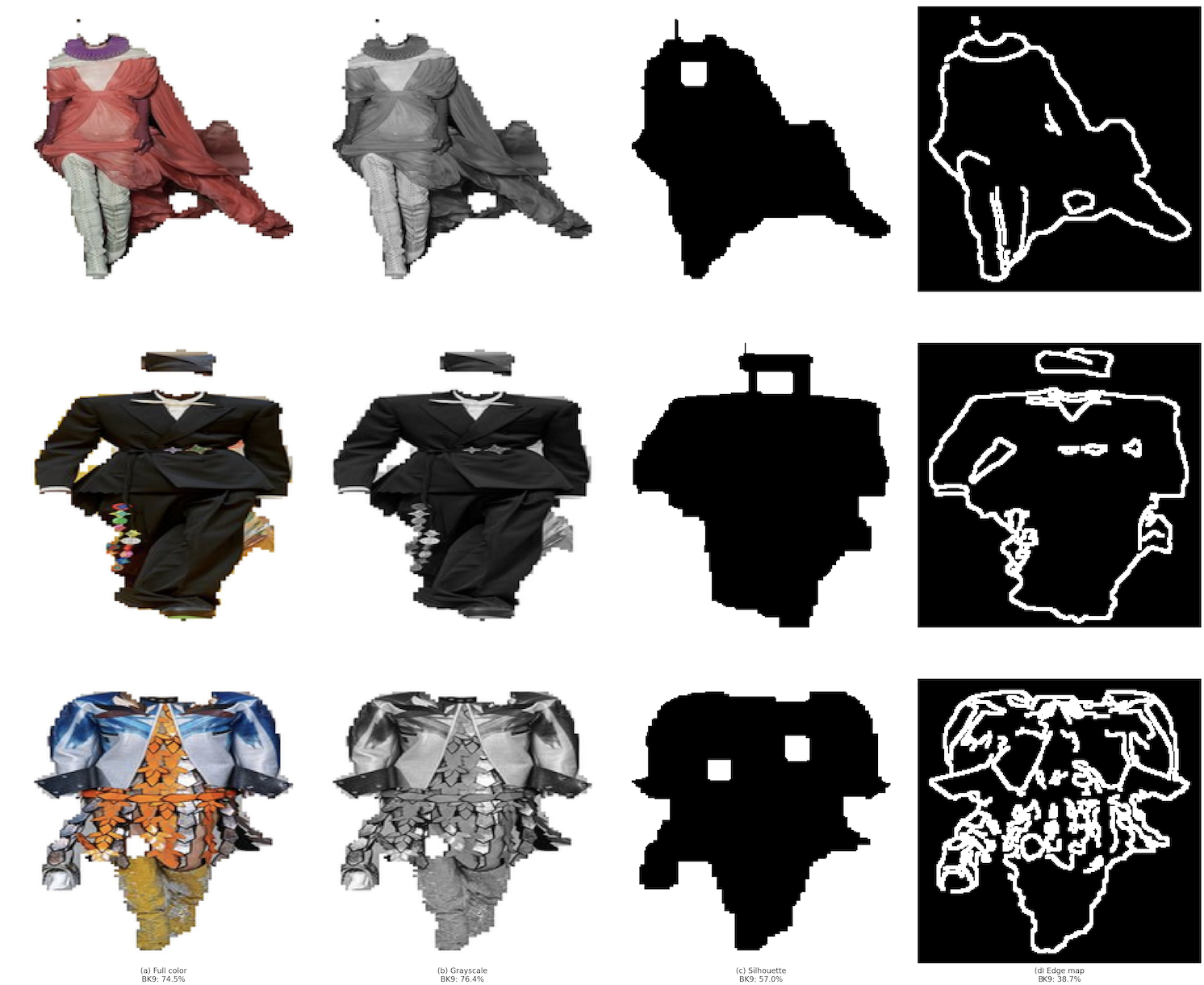}
\caption{Four representations of the same garment crop at increasing levels of abstraction. (a) Full color: hue, texture, and shape intact. (b) Grayscale: hue removed, luminance and texture retained. (c) Silhouette: surface detail removed, shape retained. (d) Edge map: contour and seam geometry only.}
\Description{Four representations of the same garment crop at increasing levels of abstraction.}
\label{fig:abstraction-ladder}
\end{figure}

\section{Dataset and Corpus}

Vogue has published continuously since 1892 and remains one of the most influential fashion media institutions globally~\cite{Vandi2023Archive}.
Unlike consumer photography or social media content, Vogue runway imagery is the product of a controlled editorial pipeline: each season, fashion houses present collections where individually cast models walk in garments chosen and styled by the house's creative team, and Vogue's editorial staff selects and publishes coverage of these presentations.
The resulting images encode a chain of deliberate aesthetic decisions, from a creative director's seasonal color strategy, through a casting director's selection of models whose appearance complements specific garments, to a stylist's final assignments~\cite{Hester2023Dress}.
When we describe this corpus as \emph{cultural} data, we mean that the associations between garment color, garment structure, and model appearance are the product of coordinated editorial judgment within Western luxury fashion rather than arbitrary pairings.
This cultural specificity is both the corpus's analytical strength, editorial logic is structured and therefore learnable, and its primary limitation: it represents one tradition, not fashion universally.

FASH-iCNN is built on 87{,}547 Vogue runway images spanning 15 fashion houses from 1991--2024.\footnote{Source: \texttt{tonyassi/vogue-runway-top15-512px} on HuggingFace.}~\cite{Guo2025VOGUE}
After quality filtering, $84{,}596$ images remain; requiring a usable face crop yields $77{,}269$.
Face crops are extracted via MediaPipe~\cite{MediaPipe2020}; clothing regions via SegFormer~\cite{Xie2021SegFormer} (ADE20K label 3), producing $65{,}541$ garment crops.
Each image is annotated with: a six-slot dominant color palette in CIELAB extracted via $k$-means on clothing pixels~\cite{Hsiao2017}, mapped to Berlin--Kay 9-class basic color terms (red, orange, yellow, green, blue, purple, pink, brown, and white)~\cite{BerlinKay1969} and CSS named colors (e.g., firebrick, goldenrod, thistle, 54 to 69 classes depending on the chromatic subset, providing finer color discrimination than Berlin--Kay families)~\cite{KayCook2023}; a Monk Skin Tone level (1--10) derived from face-crop CIELAB values~\cite{Monk2019SkinTone}; and designer, season, and year metadata.
The prediction target for the production model is the dominant slot ($c_1$) only; slots 2--6 are retained in the annotation and used in the palette-level experiments reported in Section~\ref{sec:palette-prediction}, but they are not the output of the deployed system.

\textbf{Chromatic filtering.}
The corpus is $68.3\%$ low-saturation (black or gray dominant); white is retained as a chromatic class because in editorial fashion it functions as a deliberate stylistic choice rather than a desaturation default, even though Berlin--Kay groups it with the achromatic terms~\cite{BerlinKay1969}. A model that always predicts ``black'' scores high but learns nothing about color. Removing the black/gray-dominant images yields a chromatic subset of ${\sim}24{,}500$ images (drawn from the $77{,}269$-image with-face-crop set) used for all color prediction experiments~\cite{Imtiaz2024}.
With this filtered corpus and its multimodal annotations in hand, we turn to the system that learns editorial color structure from them.

\begin{table}[!tb]
\caption{Hierarchical pipeline comparison. Each row adds one stage of constraint. The oracle row shows performance with perfect upstream predictions, establishing the pipeline's ceiling.}
\label{tab:pipeline-comparison}
\small
\centering
\begin{tabular}{@{}lcc@{}}
\toprule
Pipeline stage & $\Delta E_{00}$ & BK Acc \\
\midrule
Unconstrained LAB & 15.0 & 53.0\% \\
CSS centroid only & 9.70 & 73.8\% \\
BK$\rightarrow$CSS$\rightarrow$LAB (pred) & 9.10 & 73.4\% \\
BK$\rightarrow$CSS$\rightarrow$LAB (oracle) & 5.74 & 81.4\% \\
\bottomrule
\end{tabular}
\end{table}

\section{Multimodal Color Prediction System}

\subsection{Architecture}

Users provide a garment photograph and optionally a face photograph; clothing regions are extracted via SegFormer~\cite{Xie2021SegFormer} and each stream is processed by an independent EfficientNet-B0 backbone~\cite{Tan2019EfficientNet,Chen2021TwoStream} ($224{\times}224$ RGB input, ImageNet1K-V1 pretrained, 1280-dim output) whose features are concatenated ($\mathbb{R}^{2560}$ if both streams are present, $\mathbb{R}^{1280}$ otherwise) and passed through a two-layer head (Linear $2560{\rightarrow}512$, ReLU, Dropout $p{=}0.3$, Linear $512{\rightarrow}C$) to produce class logits. We train with AdamW~\cite{Loshchilov2019AdamW} (backbone LR $1{\times}10^{-4}$, head LR $1{\times}10^{-3}$, weight decay $1{\times}10^{-3}$), cross-entropy with label smoothing $0.1$~\cite{Muller2019LabelSmoothing}, mixed precision, ReduceLROnPlateau, and early stopping (validation-loss patience $15$, max $100$ epochs, batch size $64$) on a single NVIDIA L40S 48\,GB GPU.

The system uses a three-stage hierarchical pipeline, Berlin--Kay family prediction, CSS named-color classification within family, and constrained LAB regression around the CSS centroid, which reduces perceptual error from $\Delta E_{00}{=}15.0$ to $9.10$ against an unconstrained baseline~\cite{HeerStone2012,Sharma2005CIEDE2000}. The constrained pipeline reduces perceptual error by 39\% over unconstrained regression (Table~\ref{tab:pipeline-comparison}). The oracle ceiling ($\Delta E_{00} = 5.74$) shows that improving upstream BK and CSS classification would yield further gains, the pipeline's error is dominated by classification mistakes cascading into the regression stage, not by the regression itself.

\subsection{Garment Appearance Encodes Editorial Identity}
\label{sec:per-designer}

Table~\ref{tab:per-designer} reports per-house color accuracy. Fashion houses maintain distinct and learnable color regimes; per-designer constrained models, trained and evaluated within a single house, reach BK9 top-1 of $93.4\%$ for Calvin Klein Collection, $91.0\%$ for Chanel, and $82.3\%$ for Alexander McQueen.

\begin{table}[!ht]
\caption{Per-designer constrained clothing-only BK9 accuracy. Models are trained and evaluated within each house's chromatic subset. The four houses shown are a curated subset of the 14 trained designer-constrained models, selected to span the absolute-accuracy range (75.95\%--93.4\%) and to contrast disciplined (high-baseline) palettes with chromatically diverse (low-baseline) palettes; full per-house results across all 14 designers are available in the supplementary material.}
\label{tab:per-designer}
\small
\centering
\begin{tabular}{lccc}
\toprule
Fashion House & BK9 Top-1 & Within-House Maj. & Lift \\
\midrule
Balenciaga & 75.95\% & 46.84\% & $+29.1$pp \\
Chanel & 91.0\% & 68.5\% & $+22.5$pp \\
Alexander McQueen & 82.3\% & 60.8\% & $+21.5$pp \\
Calvin Klein Collection & 93.4\% & 80.2\% & $+13.2$pp \\
\bottomrule
\end{tabular}
\end{table}

Sorted by lift over within-house majority baseline, Balenciaga shows the strongest improvement despite lower absolute accuracy, reflecting a more chromatically diverse color regime. Calvin Klein Collection achieves the highest absolute accuracy against a high majority baseline, reflecting a disciplined achromatic palette~\cite{ZhouX2025LuxuryColor}. The distinction matters for deployment: high lift indicates genuine learning of palette variation rather than majority-class prediction. Table~\ref{tab:per-designer} shows a representative subset of 4 of the 14 trained designer-constrained models; full per-house results are available in the supplementary material.

To isolate which visual channels carry house identity, we trained
independent EfficientNet-B0 models on four abstraction levels of the
clothing crop for 14-way designer classification
(Table~\ref{tab:abstraction-unified}; one of the 15 corpus houses,
Armani Priv\'e, was excluded from this classifier for insufficient
post-filter samples).

Full-color garment appearance identifies the house at $78.2\%$ top-1,
nearly $8.5\times$ the majority baseline. Removing color while retaining
luminance and texture costs only $10.6$pp, indicating color contributes
a modest share of house identity signal. The sharpest drop occurs between
grayscale and silhouette ($-37.6$pp): texture and luminance are the
primary carriers of house identity, not shape
alone~\cite{Indrie2025Silhouettes,Yang2019Contour}. Edge map and
silhouette perform nearly identically ($30.7\%$ vs $30.0\%$), confirming
that filled shape adds little beyond contour.
The contrast is visible in Table~\ref{tab:abstraction-unified}:
the two tasks respond oppositely to information removal, revealing
that house identity and color prediction draw on different visual
channels within the same garment image.

Two experiments establish that garment appearance alone encodes temporal
editorial identity at multiple resolutions. A clothing-only EfficientNet-B0
trained on decade classification (four classes: 1991--2000, 2001--2010,
2011--2020, 2021--2024) reaches $88.6\%$ top-1 against a $45.2\%$ majority
baseline. Extending to fine-grained year prediction (34-class, 1991--2024),
the same architecture reaches $58.3\%$ top-1 against a $2.9\%$ random
baseline, lands within two years of the correct answer $73.2\%$ of the time,
and achieves a mean absolute error of $2.2$ years across 34 years (1991--2024).
House identity, decade, and specific year are therefore all recoverable from
the clothing crop alone without any metadata.

\subsection{Visual Abstraction Analysis}
\label{sec:visual-abstraction}

Fig.~\ref{fig:abstraction-ladder} illustrates the four processing stages. We test what color prediction signal is recoverable at each stage by training independent CNN models on each representation.

\begin{table}[!ht]
\caption{Visual abstraction analysis across two tasks: color prediction
(BK9, with and without face input) and designer identity prediction
(14-way). Each row is an independently trained EfficientNet-B0. ``Base''
is the BK9 majority-class baseline for the color task at that abstraction
level; ``Top-1'' is the 14-way designer-classification accuracy from a
separately trained head on the same abstraction (majority baseline
$9.3\%$). The same four abstraction levels reveal opposite compensation
patterns: face input helps color prediction most when garment information
is sparse, while designer identity collapses when texture is removed
regardless of face input.}
\label{tab:abstraction-unified}
\small
\centering
\begin{tabular}{@{}lcccc|c@{}}
\toprule
 & \multicolumn{4}{c|}{Color (BK9)} & Designer \\
Stage & Solo & +Face & Gain & Base & Top-1 \\
\midrule
Full color  & 74.5 & 73.9 & $-0.6$ & 26.8 & 78.2 \\
Grayscale   & 76.4 & 85.6 & $+9.2$ & 54.9 & 67.6 \\
Silhouette  & 57.0 & 77.8 & $+20.8$ & 52.9 & 30.0 \\
Edge map    & 38.7 & 59.2 & $+20.5$ & 26.8 & 30.7 \\
\bottomrule
\end{tabular}
\end{table}

Face input adds negligible signal when the garment stream is full-color ($-0.6$pp), but lifts accuracy by $+9.2$pp on grayscale, $+20.8$pp on silhouette, and $+20.5$pp on edge-map representations. The face input's contribution is inversely proportional to garment information richness: the system automatically derives more value from the optional face input precisely when the primary garment input is most information-poor~\cite{Ma2022Robust}.

\subsection{Modality Redundancy Analysis}
\label{sec:modality-redundancy}

\textbf{Swatch equivalence.} Replacing the full clothing crop with a flat-color swatch of its dominant color drops CSS top-1 by only $0.5$pp ($0.5254$ vs.\ $0.5302$). The garment-stream color signal at the CSS prediction level is therefore almost entirely dominant color; garment structure contributes minimally to color prediction itself.

\textbf{Face-to-designer implicit encoding.} A face-only CNN identifies the fashion house at $96.6\%$ top-1 on random splits, almost certainly inflated by subject-identity leakage; temporal splits produce substantially lower figures~\cite{Cherepanova2023,Robinson2023Faces}. Adding an explicit designer embedding to the face stream moves BK9 accuracy by only $+0.2$pp, confirming that the face modality already implicitly encodes the casting patterns the metadata would supply.

\subsection{Single-Color vs.\ Multi-Slot Prediction}
\label{sec:palette-prediction}

The production system predicts a single dominant color ($c_1$). Because the dataset annotations include all six palette slots, we ran palette-level experiments to test whether richer multi-color outputs were learnable. The results were uniformly weak, and we report them here so the production design is properly contextualized.

\textbf{Per-slot CSS prediction.} An independent CNN trained to predict the CSS class for each of the six palette slots from the clothing crop alone (no anchor) shows a sharp signal collapse beyond the dominant slot. Table~\ref{tab:per-slot} reports per-slot top-1 accuracy and median CIEDE2000 error.

\begin{table}[!ht]
\caption{Per-slot CSS prediction from clothing crop ($N=2{,}158$). The dominant slot ($c_1$) is the only slot whose median $\Delta E_{00}$ falls in perceptually similar territory; later slots degrade rapidly.}
\label{tab:per-slot}
\small
\centering
\begin{tabular}{lccc}
\toprule
Slot & \#cls & Top-1 & $\Delta E_{00}$ median \\
\midrule
$c_1$ & 68 & 0.4453 & 3.09 \\
$c_2$ & 62 & 0.3855 & 5.39 \\
$c_3$ & 53 & 0.3846 & 10.38 \\
$c_4$ & 49 & 0.3346 & 16.77 \\
$c_5$ & 45 & 0.3072 & 20.66 \\
$c_6$ & 47 & 0.3387 & 19.48 \\
\bottomrule
\end{tabular}
\end{table}

The interpretation is direct: secondary palette colors are essentially uncorrelated with the dominant-color signal the model can extract from the clothing crop. By slot 4 the median perceptual error has grown to ${\sim}17$ $\Delta E_{00}$, well outside any reasonable color-matching tolerance.

\textbf{Multi-label set prediction.} A separate CNN reframes the task as multi-label CSS classification~\cite{Zhuang2018MultiLabel}: predict the \emph{set} of CSS colors present anywhere in the six-slot palette (91 classes, BCE loss, $N=2{,}185$). On the clothing crop the model achieves precision@1 of $0.858$, precision@3 of $0.734$, precision@5 of $0.634$, macro-F1 of $0.405$, and micro-F1 of $0.652$. The precision@1 figure is the highest CSS single-color accuracy the system achieves on the full chromatic subset, but precision degrades sharply with $k$ and ordering information is lost, the model knows which colors are present but not which is dominant or how the palette is structured.

\textbf{Anchor-conditioned completion.} A third experiment supplied the model with the dominant color $c_1$ as an anchor and asked it to predict the CSS class of each subsequent slot. The anchor lifted slot-2 top-1 by $4.6$pp over a face-only baseline but the benefit faded to zero by slot 5, consistent with the per-slot finding that secondary slots are largely independent of the dominant.

Multi-color palette prediction remains an open problem on this corpus.

\section{Discussion}

\subsection{Interaction Implications}

Users engage with editorial fashion information at different levels of specificity. A user curious about the broad cultural tradition a garment belongs to can inspect the predicted house and decade and the named color tradition they represent. A user interested in how the garment fits into a color lineage can follow the named color output from Berlin--Kay family down to CSS named hue. A user making a precise styling or design decision can use the CIELAB coordinate. This layered output, from cultural provenance to perceptual color coordinate, would not be available from a flat classification or a behavioral recommender that returns similar products without disclosing the editorial logic behind them. The restriction to a single dominant color is an honest design constraint reflecting the learnability boundary established in Section~\ref{sec:palette-prediction}: the system surfaces what the signal actually supports rather than fabricating richer outputs.

\subsection{Portability and Future Work}

The pipeline framework is portable: retraining on non-Western fashion archives or regional dress collections would produce culturally distinct models with the same technical structure~\cite{Ling2019Critical,Deng2023Ethnic}. The most immediate extension is corpus diversification beyond Vogue's Western, luxury-centric coverage; a longer-term direction is understanding how cultural transparency affects user decision-making in practice~\cite{Rezwana2023COFI,Liu2025CoCreation}.

\subsection{Limitations}

Per-designer constrained color models are trained and evaluated within a single house, so cross-house color generalization is untested. All face-conditioned results should be read with the identity-leakage caveat from Section~\ref{sec:modality-redundancy} in mind, as temporal splits produce substantially lower figures than the $96.6\%$ random-split result. The pipeline's $\Delta E_{00}$ of $9.10$ is directionally meaningful but above the conventional perceptual-accuracy threshold, and palette-level prediction beyond the dominant slot remains an open problem. All evaluation is on held-out Vogue runway data; non-editorial, non-luxury, and non-Western fashion contexts are untested. Skin-tone association with garment color is negligible in the post-2000 corpus (Cram\'er's $V < 0.07$)~\cite{Mahesani2025,Nurapipah2025SkinTone}, but this observation is specific to Vogue and does not generalize.

\section{Conclusion}

FASH-iCNN demonstrates that garment appearance is a culturally structured signal from which house identity, temporal era, and color regime are independently recoverable, and that making this structure visible is a viable design principle for multimodal fashion systems. The resulting system makes its cultural reference frame inspectable rather than invisible, grounding every output in a specific, nameable editorial tradition.

\bibliographystyle{ACM-Reference-Format}
\bibliography{references}

\end{document}